\PassOptionsToPackage{table}{xcolor}
\documentclass[sigconf,9pt]{acmart}
\setcopyright{none}
\renewcommand\footnotetextcopyrightpermission[1]{}
\acmConference[]{}{}{}
\acmBooktitle{}

\usepackage{graphics}
\usepackage{booktabs}
\usepackage{graphicx}
\usepackage{subcaption}
\expandafter\def\csname ver@subfig.sty\endcsname{}
\usepackage{svg}
\usepackage{hyperref}
\usepackage{amsmath}
\usepackage{multirow}
\usepackage{colortbl}
\usepackage{lipsum}
\usepackage{enumitem}
\usepackage[super]{nth}
\usepackage[normalem]{ulem}

\usepackage{algorithm}
\usepackage{algpseudocode}
\algtext*{EndWhile}
\algtext*{EndIf}
\algtext*{EndFor}
\algtext*{EndProcedure}

\usepackage{listings}
\lstdefinestyle{jsonsmall}{
  basicstyle=\ttfamily\scriptsize,
  breaklines=true,
  frame=single,
  rulecolor=\color{gray},
  backgroundcolor=\color{white},
  columns=fullflexible,
  keepspaces=true,
  aboveskip=2pt,
  belowskip=2pt,
}

\definecolor{mygreen}{RGB}{79,173,91}

\makeatletter
\def\@mkbibcitation{\relax}
\makeatother

\DeclareUnicodeCharacter{FF0C}{,}
\DeclareUnicodeCharacter{FF08}{(}
\DeclareUnicodeCharacter{FF09}{)}
\DeclareUnicodeCharacter{FF5E}{\textasciitilde} 

\begin{document}


\title{\emph{SCALE}: \underline{S}elf-Supervised \underline{C}onstraint-\underline{A}ware \underline{L}ayout G\underline{E}neration for Local P\&R DRV Fixing at Advanced Nodes}


\author{Chia-Tung Ho$^{1,*}$, Haoyu Yang$^{1,*}$, Guanglei Zhou$^{1,2,*}$, Yoshi Nishi$^{1}$, Yaguang Li$^{1}$, Walker Turner$^{1}$, Cunxi Yu$^{1}$, Yiran Chen$^{2}$ and Brucek Khailany$^{1}$}

\affiliation{
  \institution{$^1$NVIDIA \quad $^2$Duke University}
  \country{} 
}
\thanks{*Equal contribution and corresponding authors: Chia-Tung Ho (chiatungh@nvidia.com), Haoyu Yang (haoyuy@nvidia.com), and Guanglei Zhou (gabzhou@nvidia.com).}

\begin{abstract}
As semiconductor manufacturing advances toward sub-2nm nodes, local place-and-route (P\&R) design-rule violation (DRV) fixing is increasingly limited by complex rule interactions, dense multi-layer routing geometries, and foundry-specific constraints. While Large Language Models (LLMs) have recently demonstrated strong capabilities in EDA scripting and documentation, their application to \emph{visual} layout understanding remains largely unexplored: diagnosing DRC violations from layout imagery demands precise geometric reasoning and foundry-specific rule knowledge absent from general-purpose VLM training.


We propose \emph{SCALE}, a framework with a self-supervised layout-generation stage for local DRV fixing at advanced nodes. Multi-layer layout geometry is serialized into structured text, and a fine-tuned language model learns to reconstruct randomly masked polygons from surrounding BEOL context alone without violation labels. At inference, natural-language rule constraints and high-temperature sampling steer generation toward diverse, violation-prone layout variants validated by an industrial signoff DRC checker, producing DRC-annotated layout--violation pairs used to fine-tune a domain-adapted DRC-VLM.
This VLM provides rule-aware geometric guidance for local DRV repair, boosting state-of-the-art agents' solve rates by +12--25\% (up to 97\%) on 100 real sub-2\,nm cases spanning enclosure, spacing, width, and color-spacing violations.

\end{abstract}

\maketitle 
\fancyhead[LE,RO]{}

\vspace{-6pt}
\section{INTRODUCTION}


Design Rule Checking (DRC) and Violation (DRV) correction have emerged as critical bottlenecks in achieving design closure at advanced process nodes.
At sub-2\,nm nodes, EUV lithography and Design-Technology Co-optimization (DTCO) have introduced hundreds of conditional, topology-dependent constraints (e.g., conditional enclosure, color-dependent spacing, discrete width quantization) whose legality depends jointly on neighboring geometries, multi-layer enclosure, and double-patterning color assignment.
Commercial signoff checkers~\cite{Calibre} only \emph{detect} such violations, leaving engineers to manually interpret and fix each one.
In-design DRC and detail-routing engines~\cite{Innovus,ICC2} fix violations automatically but rely on hand-tuned geometric templates~\cite{DRC+}; as rules grow increasingly conditional, this rule-by-rule approach cannot keep pace, leaving complex rules uncovered and local repairs prone to cascading new violations.
A more flexible, automated approach to local Place-and-Route (P\&R) DRV fixing is therefore needed.

\begin{figure}[t]
    \centering
    \includegraphics[width=0.425\textwidth]{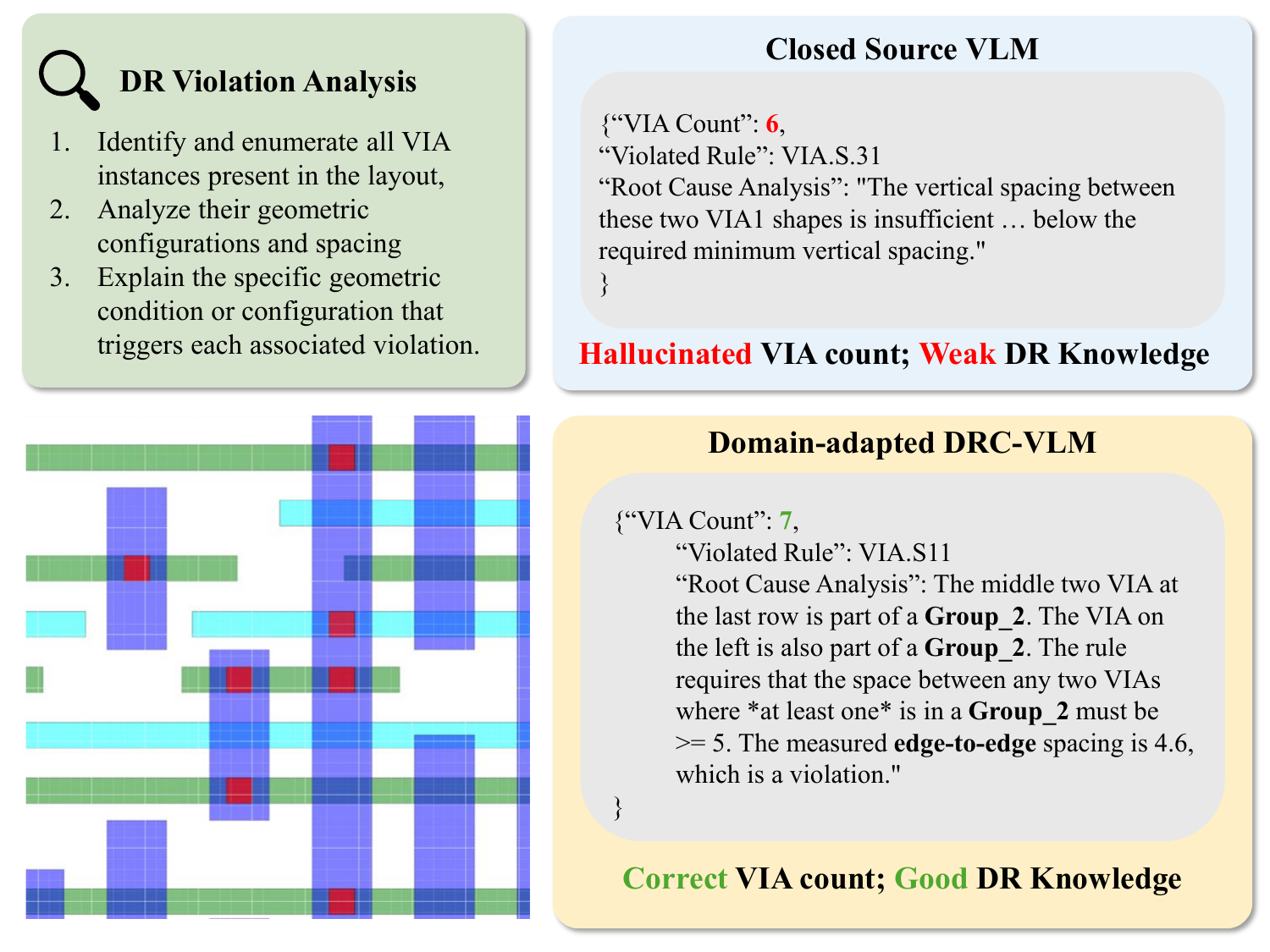}
    \caption{Off-the-shelf closed-source VLMs hallucinate geometric features (e.g., incorrect VIA count) and lack foundry-specific design rule knowledge, producing inaccurate violation analyses. Our domain-adapted VLM correctly perceives layout geometry and applies precise rule semantics, enabling reliable DRV root cause analysis.}
    \Description{Comparison of VLM outputs on a layout image: a commercial VLM hallucinates geometry while our domain-adapted VLM produces correct violation analysis.}
    \vspace{-10pt}
    \label{fig:motivation}
\end{figure}

Large Language Models (LLMs) offer a promising alternative: since design rules are textual, an LLM agent can interpret a rule constraint and propose a targeted fix without hard-coding it into a routing engine.
Prior EDA work applies LLMs to text-centric tasks such as documentation retrieval~\cite{ORAssistant}, DRC checker code generation~\cite{DRCCoder,kim2026rule2drc}, and script automation~\cite{DocumentQA_ICCAD}, but DRV fixing additionally requires visually grounding a rule in the local layout, a capability largely unexplored for vision-language models (VLMs).
As Fig.~\ref{fig:motivation} shows, off-the-shelf VLMs fail at this grounding: they hallucinate geometric features and lack the domain knowledge to map visual patterns onto rule semantics, causing agents that rely on them to misfix or reintroduce violations.
No structured layout--violation dataset exists to teach this grounding, making fine-tuning-based domain adaptation impractical.

Constructing such training data faces two barriers.
First, \emph{data scarcity}: production layouts are predominantly DRC-clean, real violations skew toward a narrow rule subset, and agentic synthesis from rule descriptions alone fails to reproduce realistic DRVs.
Second, a \emph{domain gap}: conditional design rules are foundry-confidential, so VLMs pretrained on public, natural-world images have never encountered nanometer-scale VLSI geometry and cannot generalize their spatial reasoning to it.
Overcoming both barriers is the core challenge this work addresses.

\begin{figure*}[t]
    \centering
    \includegraphics[width=0.925\textwidth]{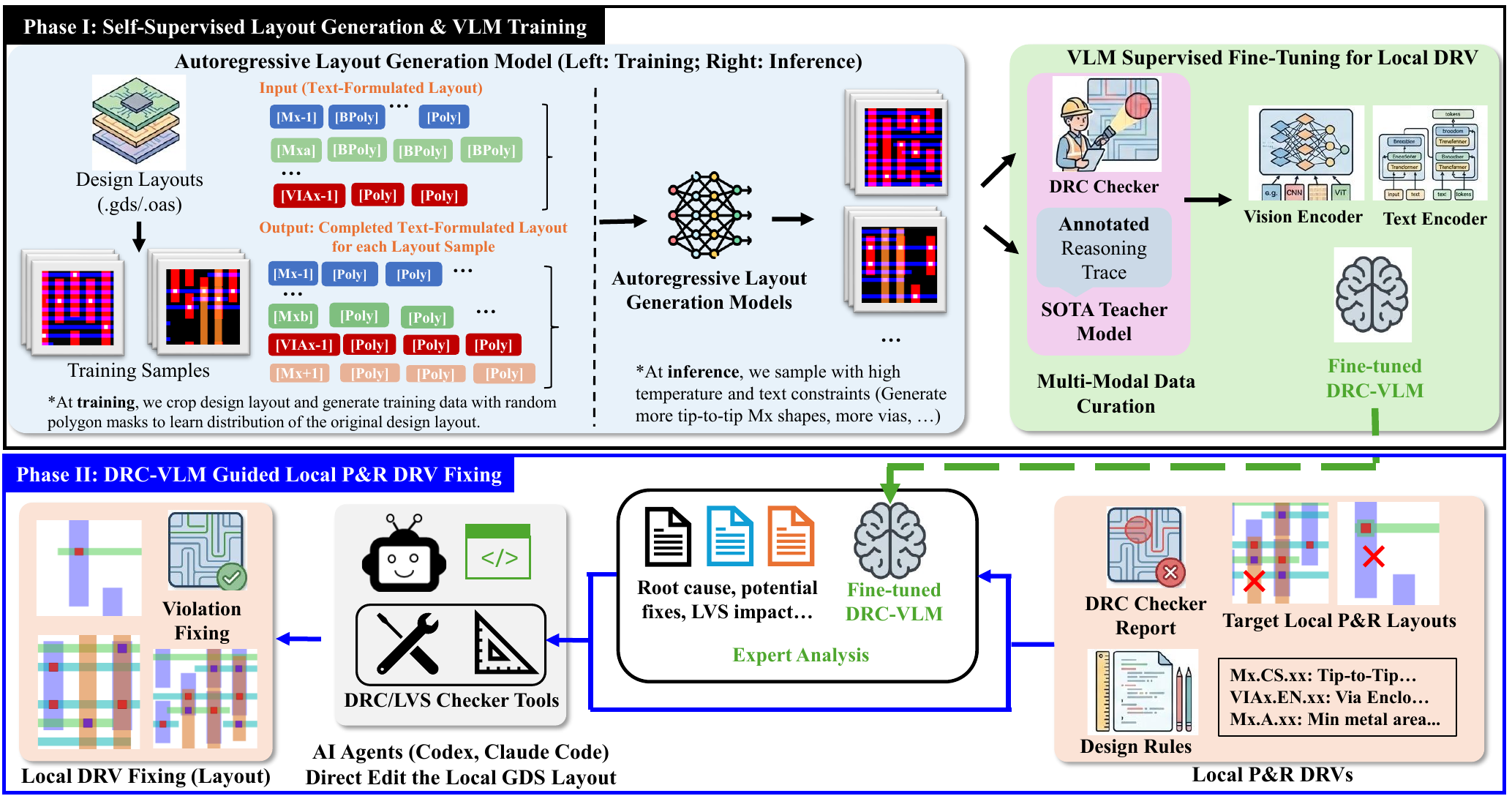}
    \caption{\textbf{Proposed Framework.} \textbf{Phase~I:} Layouts are serialized into text to train an autoregressive model; text-conditioned sampling generates diverse layouts, annotated by industrial DRC and a teacher model for fine-tuning the DRC-VLM. \textbf{Phase~II:} The DRC-VLM guides AI agents to edit the local GDS (boundary polygons may only be extended, not removed) with DRC/LVS checker-in-the-loop feedback.}
    \label{fig:proposed_framework}
    \vspace{-12pt}
\end{figure*}
Here, we address both barriers with a pipeline whose self-supervised generation stage produces diverse layout--violation pairs at scale from real designs, without hard-coding target rules into a routing/Engineering Change Order (ECO) engine or creating the DRV layouts manually.
We then fine-tune an open-source VLM~\cite{Qwen3VL} on the resulting corpus, yielding a rule-aware, domain-adapted DRC-VLM that guides a state-of-the-art~(SOTA) coding agent to fix local P\&R DRVs more efficiently and effectively than unguided agents alone, addressing violations that commercial signoff checkers can only detect but not repair.
The main contributions of this work are as follows:

\begin{itemize}[nosep,topsep=2pt,leftmargin=*]

    \item \textbf{Self-Supervised, Rule-Conditioned DRV Data Generation from Real Layouts.}
    We introduce \emph{SCALE} and develop an LLM-based generation pipeline that reconstructs masked polygons from BEOL context alone without violation labels. At inference, natural-language constraints steer sampling toward violation-prone configurations, and an industrial signoff checker verifies the outputs, yielding diverse layout--violation pairs at scale from real designs.

    \item \textbf{Constraint-Aware Data Curation and Domain-Adapted DRV Fixing.}
    On top of the generated corpus, we curate multi-category supervision (layout descriptions, violation labels, and step-by-step reasoning traces) to fine-tune a domain-adapted DRC-VLM. This model provides dynamic, rule-conditioned guidance to SOTA coding agents for local P\&R DRV repair.

    \item \textbf{Empirical Validation and SOTA Local P\&R DRV Fixing at Sub-2\,nm.} 
    We conduct holistic studies on our data generation against discrete-diffusion and agentic baselines, demonstrating superior topology diversity and rule coverage. On 100 real sub-2\,nm local P\&R fixing scenarios, DRC-VLM guidance boosts SOTA agents' solve rates by +12--25\% (up to 97\%) across enclosure, width, spacing, color-spacing, and area violations.

\end{itemize}

\vspace{-10pt}
\section{Related Work}

\noindent\textbf{LLMs and VLMs for EDA.}
LLMs have been widely adopted for text- and code-centric EDA tasks, including document and log QA~\cite{ORAssistant, DocumentQA_ICCAD, wan2025schemacoder}, chip design assistance~\cite{liu2023chipnemo}, RTL generation~\cite{ho2025verilogcoder, ScaleRTL, CraftRTL}, formal verification~\cite{wan2026fvrulelearner}, and multi-task reasoning~\cite{ho2026polymath}.
For design rule understanding, DRC-Coder~\cite{DRCCoder} and Rule2DRC~\cite{kim2026rule2drc} combine VLMs and LLMs to generate DRC checker code from rule descriptions before any violation occurs, whereas to our knowledge no prior work uses a VLM to visually ground a flagged violation, diagnose its root cause, and guide an agent to repair it.

\noindent\textbf{Spatial Reasoning in VLMs.}
General-purpose VLMs struggle with spatial reasoning~\cite{SpatialVLM, iVISPAR}, especially for VLSI layouts governed by proprietary nanometer-precision rules, motivating domain-adapted training on DRC-annotated layout--violation data.

\noindent\textbf{Layout Pattern Generation.}
Prior CNN~\cite{Deepattern}, GAN~\cite{CUP, CUP-EUV}, token~\cite{LayouTransformer}, and diffusion~\cite{Diffpattern, ControLayout, PatternPaint, Vario, DiffLayoutDFM} methods advance DFM pattern diversity~\cite{VTS, VIPER, deesign_rule_manual, OPC_Pat_gen} but are limited to single-layer rasterized images. Our stage serializes multi-layer geometry into text, trains via self-supervised masked prediction from BEOL context, and steers generation with natural-language constraints at inference, capturing cross-layer diversity.

In summary, these gaps motivate our framework: a self-supervised layout-generation stage producing DRC-annotated multi-layer layout--violation pairs, coupled with VLM fine-tuning that provides actionable guidance for local P\&R DRV fixing.

\vspace{-10pt}
\section{Methodology}
We propose a two-phase framework (Fig.~\ref{fig:proposed_framework}) that aligns VLMs with VLSI geometry to enable automated local P\&R DRV fixing. Section~3.1 describes our self-supervised data generation pipeline and VLM supervised fine-tuning for a domain-adapted DRC model (DRC-VLM); Section~3.2 describes the DRC-VLM-guided agentic framework for local P\&R DRV fixing; and Section~3.3 discusses why the two phases use different model modalities.

\vspace{-6pt}

\subsection{Phase~I: Self-Supervised Layout Generation \& VLM Supervised Fine-Tuning}

Phase~I is designed to close the data-scarcity and domain-gap barriers in two steps. Section~\ref{sec:layout_gen} trains an autoregressive model to generate diverse, DRC-annotated layout--violation pairs from real designs. Section~\ref{sec:vlm_sft} curates multi-modal supervision from this corpus and uses it to fine-tune a domain-adapted DRC-VLM.

\subsubsection{Autoregressive Layout Generation Model}
\label{sec:layout_gen}
We design the pipeline to operate on production GDS files through four stages: (1)~tile the design into crops and rasterize each layer; (2)~extract connected components and serialize into a text representation; (3)~generate target-layer \texttt{[Poly]} sequences conditioned on context and optional text constraints; (4)~decode generated tokens into physical polygons, insert into the original layout, and run industrial DRC for rule annotation. This produces a DRC-annotated, rule-labeled corpus anchored in real routed-layout topology.

\noindent\textbf{Text-Formulated Layout Representation.}
We propose a text-formulated, multi-layer layout representation (Fig.~\ref{fig:proposed_framework}) as the key enabler. Inspired by LayouTransformer~\cite{LayouTransformer}, we serialize each layout crop into structured tokens preserving layer identity, absolute position, and polygon extent. Each rectangle is encoded as a directional edge walk from its top-left vertex $(x,y)$ via cardinal deltas ($\rightarrow$~width, $\downarrow$~height, $\leftarrow$~width, $\uparrow$~height). The representation uses:
\begin{itemize}[nosep,topsep=0pt,leftmargin=*]
  \item \textbf{Layer headers} \texttt{[M1]}, \texttt{[M2]}, \texttt{[V1]} etc.: delimit each metal or via layer section.
  \item \textbf{\texttt{[Poly]}~\ldots~\texttt{[\textbackslash Poly]}}: a complete polygon fully contained within the crop window.
  \item \textbf{\texttt{[BPoly]}~\ldots~\texttt{[\textbackslash BPoly]}} (Boundary Polygon): a shape clipped by the crop boundary, used only as context, never as a generation target.
  \item \textbf{Coordinate tokens} $(x,\,y)$: absolute top-left vertex position in layout units.
  \item \textbf{Direction deltas} $\rightarrow w$, $\downarrow h$, $\leftarrow w$, $\uparrow h$: cardinal steps tracing the rectangle boundary.
\end{itemize}
This encoding is compatible with standard language-model training while preserving geometry needed for signoff-quality DRC annotation. We introduce the boundary-aware \texttt{[BPoly]} token to separate complete targets from context-only boundary shapes, ensuring the model does not treat clipped boundary shapes as valid generation targets or hallucinate geometry at crop edges.

\noindent\textbf{Self-Supervised Training \& Controllable Sampling.}
Let $X=\mathrm{Tok}(\mathcal{L}_{\mathrm{ctx}})$ denote the serialized context (both \texttt{[Poly]} and \texttt{[BPoly]}) and $Y=(y_1,\ldots,y_T)=\mathrm{Tok}(\mathcal{L}_{\mathrm{gen}})$ the generated layer sequence (only \texttt{[Poly]}). During training, the model learns to reconstruct masked polygons from context alone:
\vspace{-6pt}
\[
p_{\theta}(Y \mid X)
= \prod_{t=1}^{T} p_{\theta}(y_t \mid y_{<t}, X).
\]
\vspace{-8pt}

\noindent Training minimizes teacher-forced negative log-likelihood:
\vspace{-6pt}
\[
\mathcal{L}_{\mathrm{SFT}}(\theta)
= - \sum_{(X,Y)\in\mathcal{D}}
  \sum_{t=1}^{T} \log p_{\theta}(y_t \mid y_{<t}, X).
\]
\vspace{-8pt}

\noindent At inference, we augment the context with an optional natural-language constraint $C$ describing the target rule family (e.g., ``generate more tip-to-tip M$x$ shapes'' or ``target VIA spacing violations'') and sample tokens with temperature $\tau$:
\vspace{-6pt}
\[
y_t \sim \operatorname{softmax}\!\left(\frac{\mathbf{z}_t}{\tau}\right),
\quad
\mathbf{z}_t=f_{\theta}(y_{<t},X,C),
\]
\vspace{-8pt}

\noindent where larger $\tau$ increases diversity and the likelihood of rare rule-triggering configurations. Constraint following relies on the instruction-following capability of the pretrained language model; layout fine-tuning teaches only the structured polygon representation, leaving the model's ability to condition on natural-language prompts intact. We decode sampled sequences back into polygons and insert them into the original layout; signoff DRC checker then annotates each candidate with the violated rule and violation location.

\begin{table}[t]
\caption{Training data composition by generation method and layer type in Section~\ref{Sec:VLMSFTExp}. The DRC analysis task is augmented with reasoning source for the proposed in Table~\ref{tab:combined_metal_via_results}. The description task includes via count, via location, and metal/via bounding box extraction tasks.}
\label{tab:data_composition}
\centering
\small
\begin{tabular}{llrrr}
\hline
\multirow{2}{*}{\textbf{Method}} &
\multirow{2}{*}{\textbf{Layer Type}} &
\multicolumn{1}{c}{\textbf{DRC Analysis}} &
\multicolumn{1}{c}{\textbf{Description}} &
\multirow{2}{*}{\textbf{Total}} \\
& & \multicolumn{1}{c}{\textbf{(w/wo Reasoning)}} &
\multicolumn{1}{c}{\textbf{Task}} & \\
\hline
\multirow{2}{*}{Proposed}
  & VIA   & 7,606 & 9,255  & 16,861 \\
  & Metal & 8,946 & 8,308  & 17,254 \\
\hline
\multirow{2}{*}{Script}
  & VIA   & 6,244 & 9,255  & 15,499 \\
  & Metal & 5,583 & 11,166 & 16,749 \\
\hline
\end{tabular}
\end{table}

\subsubsection{VLM Supervised Fine-Tuning for Local DRV}
\label{sec:vlm_sft}
We run a DRC checker~\cite{Calibre} on the generated layout crops from Section~\ref{sec:layout_gen} to obtain ground-truth violation labels and locations, which we then curate into two complementary multi-modality supervision categories (Table~\ref{tab:data_composition}):
\begin{itemize}[nosep,topsep=2pt,leftmargin=*]
\item \textbf{DRC Analysis Task:} A teacher model
  (Gemini-3-Pro) receives each layout crop image together with its
  DRC violation report and generates a reasoning trace that
  identifies relevant layout features, measures dimensions, compares
  against rule thresholds, and concludes with the violated rule. We
  apply multi-stage filtering on rule coverage, violation polygon
  location, and direction/spacing consistency, removing
  13.40\% of traces with image input
  (Table~\ref{tab:filter_stage}).
\item \textbf{Description Task:} We pair each crop with programmatically extracted metal shape and VIA center coordinates in a structured JSON response, teaching the model to perceive, count, and localize geometric features.
\end{itemize}

We combine both data categories into a unified, shuffled training mixture for supervised fine-tuning of Qwen3-VL-8B~\cite{Qwen3VL}. Each sample pairs a layout crop and rule description with either a structured JSON prediction or a reasoning trace followed by the prediction. We train the model with autoregressive cross-entropy loss over target tokens (prompt and image tokens masked), jointly optimizing layout perception, violation classification, and diagnostic reasoning.

\vspace{-4pt}
\subsection{Phase II: VLM-Guided Local P\&R DRV Fixing}

We employ the domain-adapted DRC-VLM as an expert diagnostic front-end to a SOTA AI agent for local P\&R DRV repair (Fig.~\ref{fig:proposed_framework}).
We input the DRC-VLM with a rendered image of the local layout region, the DRC error report, and the relevant design-rule description; it produces a structured expert analysis: root-cause identification, the specific polygons involved with their coordinates, and a ranked list of top-$k$ potential fix actions with predicted connectivity impact.

This analysis is then provided to a SOTA AI coding agent equipped with DRC and LVS checker tools.
The agent directly edits the local GDS without a predefined action space: it may move, resize, add, or remove interior polygons as needed, while boundary polygons may only be extended or supplemented, not removed, to preserve connectivity with the surrounding design.
DRC and LVS checkers are invoked after every edit to verify rule legality and preserve circuit connectivity, forming a checker-in-the-loop repair cycle.
Unlike template-based engines that narrow the search space through static, pre-encoded correction patterns, the DRC-VLM provides dynamic, rule-conditioned guidance grounded in domain knowledge of complex conditional design rules, enabling the agent to adapt its fixes to each violation's geometric and topological context.

\subsection{Discussion: Choosing the Model Modality}
\label{Sec:ModalityDiscuss}


We use a text LLM for layout augmentation because our structured
representation enables diverse outputs that decode directly into
editable polygons, whereas image-based generation would require
vectorization, grid snapping, layer assignment, and filtering before
signoff. DRC reasoning, however, demands understanding of spatial
relations such as spacing, enclosure, and multi-layer interactions
that layout images expose directly but polygon sequences flatten into
long token streams. To validate this, we compare the filtering rates
of reasoning traces generated from text-represented layouts versus
image inputs on the DRC Analysis Task
(Table~\ref{tab:filter_stage}). 
Text input leads to a 30.66\%
overall filtering rate versus only 13.40\% for image input, with the
gap concentrated at the location stage (21.60\% vs.\ 3.50\%): the
text representation enables correct rule identification and logically
valid reasoning, yet lacks the spatial context to distinguish the
exact violating polygon from geometrically identical neighbors. These
results confirm complementary roles for the two modalities, text for
controllable layout generation and vision for spatial DRC analysis,
and demonstrate that visual input is irreplaceable for polygon-level
localization in dense layouts.

\begin{table}[h]
\centering
\caption{Filtering rate of reasoning traces at each stage for text-represented layout versus image input on the DRC Analysis Task.}
\label{tab:filter_stage}
\small
\begin{tabular}{lcccc}
\hline
 & \multicolumn{4}{c}{Filter Stage} \\
\cline{2-5}
 & DRV Rule & Location & Dir./Measure & Total \\
\hline
Text Rep. & 4.03\% & 21.60\% & 5.03\% & 30.66\% \\
Image     & 9.60\% & 3.50\%  & 0.30\% & 13.40\% \\
\hline
\end{tabular}
\end{table}
\vspace{-6pt}
\section{Experimental Results}\label{sec:experiments}
Our evaluation covers layout data generation quality, DRV detection and understanding, and local DRV fixing at an advanced technology node.
Both the auto-regressive layout generation model (Qwen2.5-3B-Instruct) and DRC-VLM (Qwen3-VL-8B) are fine-tuned using AdamW ($\beta_1$\,=\,0.9, $\beta_2$\,=\,0.95) with a cosine learning-rate schedule on 4$\times$ A100-80GB GPUs.
For the DRV rule set, we select a subset of local P\&R relevant DRC rules from the thousands available at the sub-2\,nm technology node, covering spacing, enclosure, color spacing, group via spacing, and width constraints.

\begin{table}[t]
\centering
\caption{Pattern diversity comparison across generation methods. $H$ is the Shannon entropy of the joint complexity distribution $P(c_x, c_y)$. Higher $H$ indicates more diverse layout patterns. DRC columns report design rule violation metrics for VIA and Metal layers separately. DRC columns report unique validated triggered rules by category (VIA / Metal). Abbreviations: W = Width, W+A = Width + Area, Spc = Spacing, G-Spc = Group-conditioned Spacing, EN = Enclosure, Grp = Group, R = Routing, CS = Colored Spacing.}
\label{tab:dataset_metrics}
\scriptsize
\setlength{\tabcolsep}{3pt}
\begin{tabular}{lcc|cccccc}
\toprule
\textbf{Method} & \textbf{H} $\uparrow$ & \textbf{Unique Top.} & \multicolumn{6}{c}{\textbf{Validated VIA / Metal DRV Type}}  \\
\cmidrule(lr){4-9} 
& &  & \textbf{W / W+A} & \textbf{Spc} &  \textbf{EN} & \textbf{G-Spc / R}  & \textbf{Grp / CS} & \textbf{Total} \\
\midrule
Training Data                  & 2.93 &  93  & -- & -- & -- & -- & -- & -- \\
D3PM                           & 4.61 & 236  & 1 / 5 & 5 / 4 & 7 / 2 & 3 / 1  & 2 / 1 & 18 / 13 \\
Script                         & 3.94 & 133  & 2 / 7 & 3 / 2 & 1 / 2 & 2 / 1  & 2 / 0 & 10 / 12 \\
Proposed                       & 5.07 & 289  & 1 / 13 & 13 / 4  & 6 / 4 & 7 / 10  & 3 / 3 & 29 / 34 \\
\bottomrule
\end{tabular}%
\end{table}
\begin{figure}[!t]
    \centering
    \includegraphics[width=0.5\textwidth]{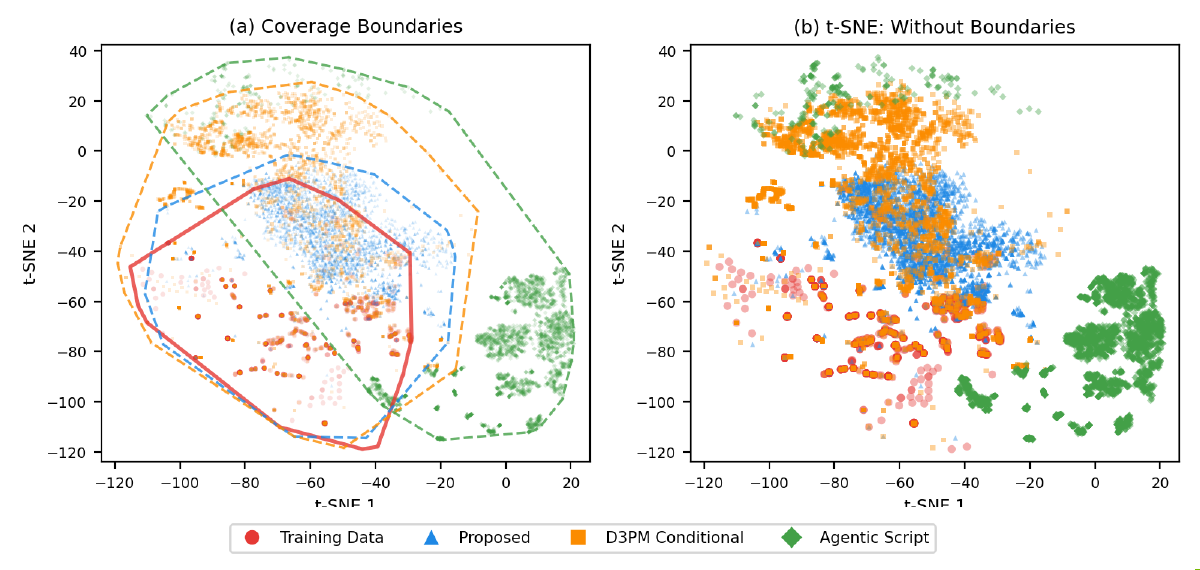}
    \caption{\textbf{t-SNE Diversity Analysis.} t-SNE projection comparing synthetic datasets against the ground truth (red). Both the trained proposed (autoregressive) and D3PM models achieve high-fidelity, in distribution coverage. In contrast, the heuristic script-based baseline fails to capture the realistic geometric distribution.}
    \Description{t-SNE scatter plots showing overlap of synthetic datasets with ground truth layout distribution.}
    \label{fig:tsne_coverage}
    \vspace{-6pt}
\end{figure}

\subsection{DRV Layout Data Generation Evaluation}

We evaluate the quality of the generated DRC violation layout dataset. The auto-regressive layout generation model is trained on 1,000 real design layout crops using random polygon masking and reconstruction from BEOL context alone, without constraints or violation labels. During inference, we supply rule-specific or geometric constraints (e.g., generate more VIA groups, more tip-to-tip M$x$ shapes) and sample under high temperature to steer generation toward violation-prone configurations.
Each method generates ${\sim}$15K samples; after filtering out-of-scope and near-duplicate DRVs, these are used for the downstream VLM training comparison in Section~\ref{Sec:VLMSFTExp}.

\noindent\textbf{Baselines.}
We compare against two baselines: a discrete diffusion model (D3PM)~\cite{PatternPaint} trained on the same data, and an agentic script-based method in which a SOTA coding agent~\cite{claudecode2026} is provided with DRC rule descriptions and layer parameters (routing directions, pitch, and width) to generate Python scripts. These scripts are then executed to produce synthetic DRV layout--violation pairs.

\noindent\textbf{Evaluation Metrics and Results.}
We measure generation quality using layout entropy ($H$)~\cite{Deepattern, LayouTransformer}, unique topologies, and validated triggered design rules, as shown in Table~\ref{tab:dataset_metrics}. Following~\cite{LayouTransformer}, $H$ is the Shannon entropy over scan-line counts $(c_x,c_y)$:
\vspace{-4pt}
\[
H = - \sum_i \sum_j P(c_{x_i}, c_{y_j}) \log P(c_{x_i}, c_{y_j}).
\]
\vspace{-6pt}

\begin{figure}[!t]
    \centering
    \includegraphics[width=0.48\textwidth]{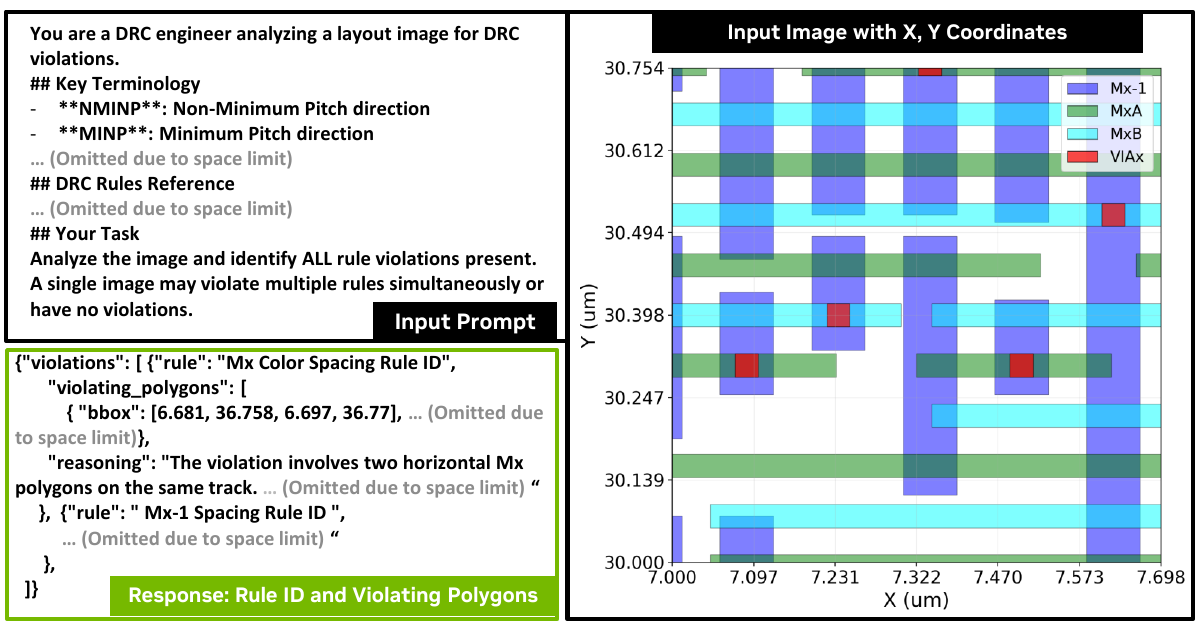}
    \caption{An example of the VLM detection task used for SFT. 
    Each SFT training pair consists of a layout image, input prompt, and response. \textcolor{blue}{Axis coordinates are scaled for confidentiality.}}
    \label{fig:sftinputexample}
    \vspace{-8pt}
\end{figure}

\noindent Unique topologies are distinct $(c_x,c_y)$ pairs; rule coverage counts validated triggered rules after filtering out-of-scope and near-duplicate DRVs.
Our method achieves the highest entropy ($H=5.07$), the most unique layouts (289), and the broadest validated rule coverage, triggering 29 VIA and 34 metal rule types. The t-SNE visualization in Fig.~\ref{fig:tsne_coverage}, further confirms that the auto-regressive generator and D3PM match the geometric distribution of real routed layouts, while the agentic script baseline produces simpler patterns that deviate from real designs.




\vspace{-10pt}
\subsection{VLM Supervised Fine-Tuning Result} \label{Sec:VLMSFTExp}

The DRV detection task is designed to strengthen the model's understanding of layout geometry and design-rule constraints as shown in Fig.~\ref{fig:sftinputexample}. Unlike simple object localization, this task requires the VLM to reason about spatial relationships between polygons, measure rule-relevant distances, and identify violation-causing regions. This provides high-value supervision for downstream layout tasks, which are diverse, scenario-dependent, and difficult to exhaustively construct as training data.

\begin{table}[!t]
\centering
\caption{DRV detection testing dataset rule frequency by category.}
\label{tab:benchmark_statistics}
\scriptsize
\begin{tabular}{lcc}
\toprule
\textbf{VIA / Metal Category} & \textbf{VIA (100 samples)} & \textbf{Metal (100 samples)} \\
\midrule
W / W+A        & 5   & 180 \\
Spc            & 177 & 108 \\
EN             & 3   & 81  \\
G-Spc / R      & 308 & 25  \\
Grp / CS       & 30  & 15  \\
G-Spc / EN     & 308 & 81  \\
\midrule
Total triggers & 523 & 409 \\
Avg./sample    & 5.23 & 4.09 \\
\bottomrule
\end{tabular}
\end{table}

\begin{table}[t]
\centering
\caption{DRV detection results averaged across Metals and VIAs. Rule F1 and Loc F1 measure rule identification and rule-gated localization accuracy; Comb F1 is their harmonic mean. All values are pass@$k$ over 10 trials. \textsuperscript{$\dagger$}Ablations isolating data source and reasoning contributions.}
\label{tab:combined_metal_via_results}
\resizebox{0.48\textwidth}{!}{%
\begin{tabular}{l ccc ccc ccc}
\toprule
& \multicolumn{3}{c}{\textbf{Rule F1 (\%)}} & \multicolumn{3}{c}{\textbf{Loc F1 (\%)}} & \multicolumn{3}{c}{\textbf{Comb F1 (\%)}} \\
\cmidrule(lr){2-4} \cmidrule(lr){5-7} \cmidrule(lr){8-10}
\textbf{Model} & @1 & @3 & @5 & @1 & @3 & @5 & @1 & @3 & @5 \\
\midrule
\multicolumn{10}{l}{\textit{Commercial \& Open-Source Baselines}} \\
Gemini-3-Pro & 34.9 & 46.2 & 49.6 & 14.4 & 21.5 & 24.6 & 20.4 & 29.3 & 32.8 \\
GPT-5.2      & 24.0 & 35.3 & 40.2 & 7.4  & 14.0 & 17.5 & 10.7 & 19.3 & 23.6 \\
Qwen3-Base   & 19.4 & 38.6 & 47.4 & 3.7  & 8.8  & 12.4 & 6.2  & 14.3 & 19.4 \\
\midrule
\multicolumn{10}{l}{\textit{Qwen3-VL-8B-SFT}} \\
\rowcolor{gray!8}
\quad Mixed-NoR-Script\textsuperscript{$\dagger$}  & 37.0 & {48.5} & {54.0} & 4.8  & 10.5 & 14.0 & 7.6  & 15.9 & 20.7 \\
\rowcolor{gray!8}
\quad Mixed-NoR-AutoReg\textsuperscript{$\dagger$}  & 17.3 & 49.5 & 55.7 & 15.1 & 25.9 & 31.2 & 20.3 & 33.5 & 39.4 \\
\rowcolor{gray!8}
\quad DRC-VLM (Proposed) & \textbf{43.8} & \textbf{57.4} & \textbf{61.6} & \textbf{23.5} & \textbf{31.9} & \textbf{35.6} & \textbf{29.5} & \textbf{39.9} & \textbf{44.2} \\
\bottomrule
\end{tabular}%
}
\end{table}

\begin{table*}[ht]
\centering
\caption{Local DRV fixing results and token cost per case. Baseline agents (Claude Code, Codex) are SOTA coding agents with access to DRC rule descriptions, raw DRC reports, rendered layout images, and DRC/LVS checker tools. Proposed methods augment each agent with expert analysis from DRC-VLM, the strongest model selected from Section~\ref{Sec:VLMSFTExp}. Solved Count (SC) and Solve Rate (SR) are reported per DRV rule category and in total. Absolute SR improvement is shown below proposed entries (\textcolor{green!60!black}{green}). Token cost in thousands (K): Codex reports \texttt{tokens\_used}; Claude Code reports output + cache-creation tokens (excluding cache-read input). Relative reduction below (\textcolor{green!60!black}{green} = fewer, \textcolor{red}{red} = more). Bold = best.}
\resizebox{\textwidth}{!}{
\begin{tabular}{l|ccccc|c|rrrrrr}
\hline
\multirow{2}{*}{Method}
& \multicolumn{5}{c|}{DRV Rule Category SC / SR (\%)}
& {Total}
& \multicolumn{6}{c}{Avg. Tokens / Case (K)} \\
\cline{2-6} \cline{8-13}
& EN & Width & Spacing & Color Spacing & Area
& SC / SR (\%)
& EN & Width & Spacing & Color Spacing & Area & Overall \\
\hline
\textbf{Baselines} & & & & & & & & & & & & \\
Claude Code
& 11 (28.2\%) & 26 (96.3\%) & 20 (95.2\%) & 5 (50.0\%) & 1 (33.3\%)
& 63 (63.0\%)
& 300.8 & 82.0 & 96.2 & 44.4 & 187.7 & 169.7 \\
Codex
& 33 (84.6\%) & 26 (96.3\%) & 18 (85.7\%) & 5 (50.0\%) & \textbf{3 (100.0\%)}
& 85 (85.0\%)
& 96.0 & 82.3 & 92.2 & \textbf{28.4} & \textbf{130.3} & 85.8 \\
\hline
\textbf{Proposed} & & & & & & & & & & & & \\
DRC-VLM + Claude Code
& \begin{tabular}[c]{@{}c@{}}32 (82.1\%)\\{\small \textcolor{green!60!black}{+53.8\%}}\end{tabular}
& \begin{tabular}[c]{@{}c@{}}\textbf{27 (100.0\%)}\\{\small \textcolor{green!60!black}{+3.7\%}}\end{tabular}
& \begin{tabular}[c]{@{}c@{}}\textbf{21 (100.0\%)}\\{\small \textcolor{green!60!black}{+4.8\%}}\end{tabular}
& \begin{tabular}[c]{@{}c@{}}6 (60.0\%)\\{\small \textcolor{green!60!black}{+10.0\%}}\end{tabular}
& \begin{tabular}[c]{@{}c@{}}2 (66.7\%)\\{\small \textcolor{green!60!black}{+33.3\%}}\end{tabular}
& \begin{tabular}[c]{@{}c@{}}88 (88.0\%)\\{\small \textcolor{green!60!black}{+25.0\%}}\end{tabular}
& \begin{tabular}[c]{@{}r@{}}214.3\\{\small \textcolor{green!60!black}{(28.7\%)}}\end{tabular}
& \begin{tabular}[c]{@{}r@{}}\textbf{74.4}\\{\small \textcolor{green!60!black}{(9.3\%)}}\end{tabular}
& \begin{tabular}[c]{@{}r@{}}\textbf{83.0}\\{\small \textcolor{green!60!black}{(13.7\%)}}\end{tabular}
& \begin{tabular}[c]{@{}r@{}}\textbf{21.5}\\{\small \textcolor{green!60!black}{(51.5\%)}}\end{tabular}
& \begin{tabular}[c]{@{}r@{}}162.4\\{\small \textcolor{green!60!black}{(13.5\%)}}\end{tabular}
& \begin{tabular}[c]{@{}r@{}}128.2\\{\small \textcolor{green!60!black}{(24.5\%)}}\end{tabular} \\
DRC-VLM + Codex
& \begin{tabular}[c]{@{}c@{}}\textbf{39 (100.0\%)}\\{\small \textcolor{green!60!black}{+15.4\%}}\end{tabular}
& \begin{tabular}[c]{@{}c@{}}\textbf{27 (100.0\%)}\\{\small \textcolor{green!60!black}{+3.7\%}}\end{tabular}
& \begin{tabular}[c]{@{}c@{}}\textbf{21 (100.0\%)}\\{\small \textcolor{green!60!black}{+14.3\%}}\end{tabular}
& \begin{tabular}[c]{@{}c@{}}\textbf{7 (70.0\%)}\\{\small \textcolor{green!60!black}{+20.0\%}}\end{tabular}
& \begin{tabular}[c]{@{}c@{}}\textbf{3 (100.0\%)}\\{\small +0.0\%}\end{tabular}
& \begin{tabular}[c]{@{}c@{}}\textbf{97 (97.0\%)}\\{\small \textcolor{green!60!black}{+12.0\%}}\end{tabular}
& \begin{tabular}[c]{@{}r@{}}\textbf{91.8}\\{\small \textcolor{green!60!black}{(4.6\%)}}\end{tabular}
& \begin{tabular}[c]{@{}r@{}}83.5\\{\small \textcolor{red}{(-1.4\%)}}\end{tabular}
& \begin{tabular}[c]{@{}r@{}}88.9\\{\small \textcolor{green!60!black}{(3.6\%)}}\end{tabular}
& \begin{tabular}[c]{@{}r@{}}29.8\\{\small \textcolor{red}{(-4.9\%)}}\end{tabular}
& \begin{tabular}[c]{@{}r@{}}\textbf{130.3}\\{\small (0.0\%)}\end{tabular}
& \begin{tabular}[c]{@{}r@{}}\textbf{83.9}\\{\small \textcolor{green!60!black}{(2.2\%)}}\end{tabular} \\
\hline
\end{tabular}
}
\label{tab:drc_combined}
\end{table*}

\noindent\textbf{Training Data Composition.}
Table~\ref{tab:data_composition} summarizes the training data. After filtering DRC violations and removing near-duplicate tasks, our proposed pipeline yields 16,861 VIA and 17,254 metal samples spanning DRC analysis (with both reasoning and non-reasoning supervision) and layout description tasks (via counting, localization, and bounding-box extraction). As a baseline, we collect 15,499 VIA and 16,749 metal samples using script-based generation with the same filtering procedures.


\noindent\textbf{DRV Detection Testing Dataset.}
We construct a 200-sample test set (100 VIA, 100 metal) from real layout crops of a \emph{different IP} than the training data; both share the same sub-2\,nm rule deck but have disjoint routing patterns. Each sample may contain multiple co-occurring violations; as shown in Table~\ref{tab:benchmark_statistics}, the set contains 523 VIA-rule and 409 metal-rule triggers (avg.\ 5.23 and 4.09 per sample) spanning categories mentioned in Table~\ref{tab:dataset_metrics}.

\noindent\textbf{Evaluation Metrics.}
Each response is scored at two levels: \emph{Rule identification} via set-level F1 between predicted and ground-truth violated-rule sets, and \emph{Localization}, gated by rule correctness—a prediction $(\hat{r},\hat{p})$ is a TP only if $\hat{r}$ matches a ground-truth rule and $\hat{p}$ matches its ground-truth location, else FP:
\vspace{-4pt}
\[
\text{Loc}(\hat{r}, \hat{p}) =
\begin{cases}
\mathrm{TP}, & \hat{r} \in \mathcal{R}_{\mathrm{GT}} \ \text{and}\  \hat{p} \leftrightarrow \mathcal{P}_{\mathrm{GT}}(\hat{r}),\\
\mathrm{FP}, & \text{otherwise},
\end{cases}
\]
\vspace{-6pt}

\noindent Unmatched ground-truth instances count as FN. Each sample is evaluated over $n=10$ trials and aggregated via the pass@$k$ estimator~\cite{chen2021evaluating} for Rule F1 and Loc F1; Comb F1 is their harmonic mean.

\noindent\textbf{Baselines.}
We compare against strong SOTA VLM baselines, including Gemini-3-Pro, GPT-5.2, and Qwen3-VL-8B-Base, using identical rule descriptions and layout images as input.

\begin{figure*}[t]
    \centering
    \vspace{-10pt}
    \includegraphics[width=0.95\textwidth]{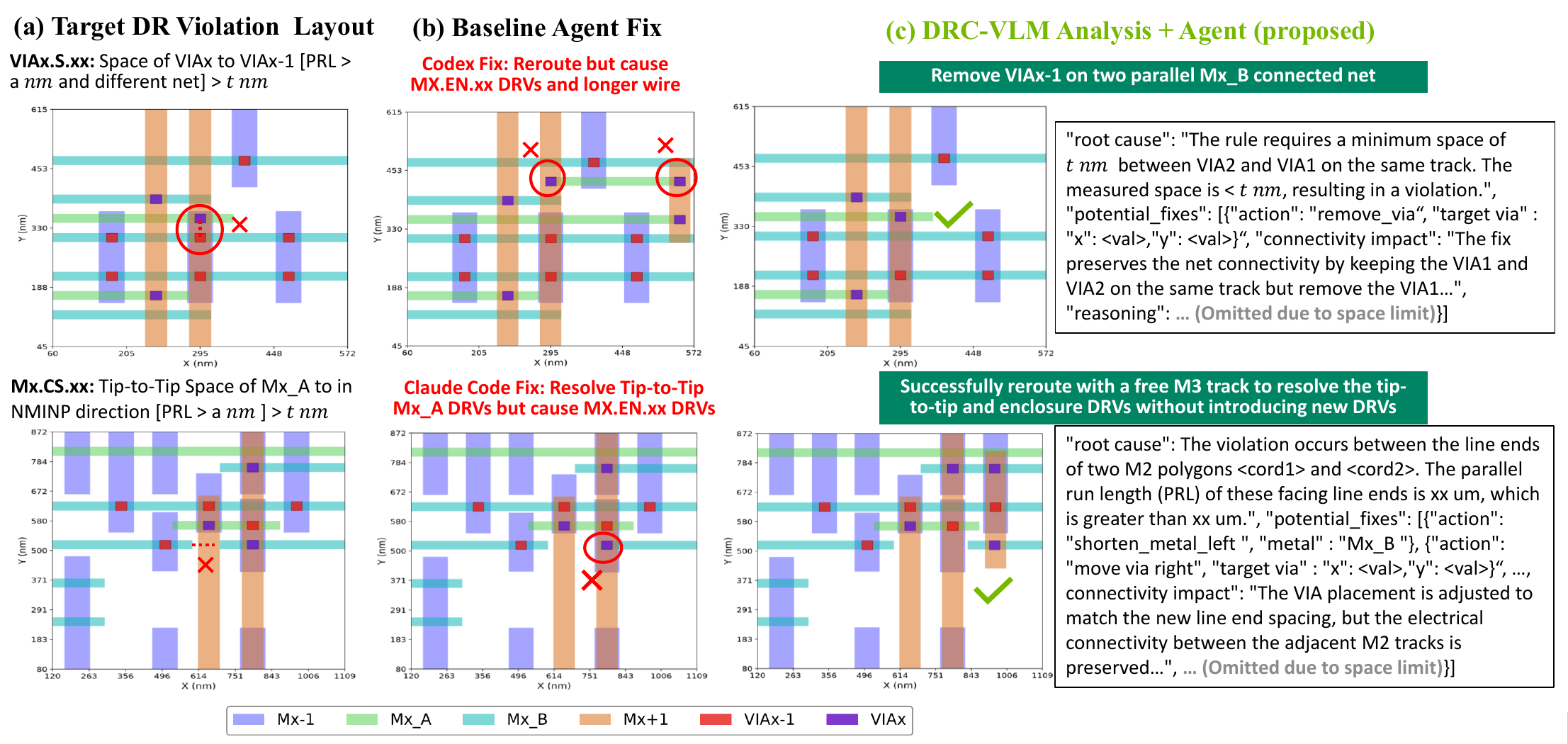}
    \caption{Examples of DRC fix approaches for VIA spacing (top) and metal tip-to-tip (bottom) rules. (a) Target layouts with the original violations. (b) Baseline agent fixes introduce new violations while try to fix the target DRV. (c) Proposed DRC-VLM analysis with structured root-cause and connectivity-aware actions yielding clean fixes. PRL=Parallel Running Length. \textcolor{blue}{Axis coordinates are scaled for confidentiality.}}
    \label{fig:case_study}
    \vspace{-10pt}
\end{figure*}

\noindent\textbf{Evaluation Result.} Table~\ref{tab:combined_metal_via_results} shows that higher-quality generated training data directly improves VLM performance on DRV detection: our method achieves the best overall pass@5 Comb F1 of 44.2, a 24.8-point gain over the base model, versus 32.8 for Gemini-3-Pro. The main bottleneck for general VLMs is localization rather than rule recall: Gemini-3-Pro reaches only 24.6 Loc F1 at pass@5, GPT-5.2 reaches 17.5, and Qwen3-VL-8B-Base reaches 12.4, despite receiving the same rule descriptions and layout images. The gain from our data is also largest in localization---the script-based baseline (Mixed-NoR-Script) attains high Rule F1 but poor Loc F1, showing it exposes rule labels without the geometric fidelity needed for spatial grounding---confirming that realistic generated layouts are critical for accurate violation localization.
This aligns with the filtering analysis in Section~\ref{Sec:ModalityDiscuss}: even Gemini-3-Pro, our teacher model, achieves only 24.6 Loc F1---it can generate reasoning traces but cannot reliably localize violating polygons, making fine-tuning on our DRC-annotated layout data essential for Phase~II guidance.
As a result, our domain-adapted DRC-VLM is used next to guide local P\&R DRV fixing.

\vspace{-6pt}
\subsection{Local DRC Fixing Experiment}
We demonstrate that DRC-VLM can significantly improve SOTA AI agents on fixing local P\&R DRC violations at sub-2nm technology node.
Given that general VLMs fail at the prerequisite localization task (Table~\ref{tab:combined_metal_via_results}), we employ our domain-adapted DRC-VLM as the expert front-end.
For each DRV, we extract a local GDS region surrounding the violation site as the repair context.
Interior polygons may be freely edited, while boundary polygons may only be extended or supplemented, not removed, to preserve surrounding connectivity.
Given the target DRV error report, the local layout netlist, and a rendered image of the region, the fine-tuned DRC-VLM analyzes the violation and proposes an expert analysis including top-$k$ potential fixes, where $k=3$.
This expert analysis is then supplied to a SOTA coding agent equipped with DRC and LVS checker tools to fix the local DRVs.
We run the agent fix once per problem for both proposed method and baselines for a fair comparison.

\noindent\textbf{Local P\&R DRV Dataset.}
From an unseen block-level IP design that was DRC-clean under a prior developing rule deck but exhibits numerous errors under the latest rule deck at sub-2nm, we curate 100 local P\&R DRV cases by selecting representative error regions and filtering duplicates. The dataset spans five categories (EN: 39, Width: 27, Spacing: 21, Color Spacing: 10, Area: 3) covering same-layer and cross-layer interactions, making it representative of local Engineering Change Order (ECO) scenarios where geometric legality, connectivity, and advanced-node context must all be satisfied.
Curation prioritizes hard-to-fix conditional rules (enclosure, color spacing); the category distribution matches the actual P\&R error distribution.
Each instance requires up to 1{,}800\,s of agent compute with iterative checker feedback, making large-scale evaluation costly; the Area category (3 cases) is included for completeness but is too small to draw per-category conclusions.

\noindent\textbf{Baselines.}
We compare against two SOTA coding agents~\cite{claudecode2026, OpenAICodex2026} \emph{without} the proposed expert analysis from DRC-VLM. Claude Code uses Sonnet~4.6~\cite{anthropic2026claude} as its core model, and Codex uses GPT-5.4~\cite{OpenAI_GPT54}. Each agent is provided with DRC rule descriptions, metal pitch and preferred routing direction parameters, the raw DRC report, the input GDS, and the rendered PNG layout image in a shared working directory. Both agents have access to DRC and LVS signoff checker tools, with an 1800\,s time limit imposed per DRV fixing instance.
These constitute strong baselines: as SOTA coding agents, they can programmatically analyze DRC reports, restructure rule and layout information, inspect rendered images, and iteratively repair the GDS with checker-in-the-loop feedback.

\noindent\textbf{Evaluation Metrics.}
A case is \emph{solved} if the target DRV is fixed without introducing new DRVs in the region \emph{and} the design passes LVS. We report the solve rate per category and overall.

\noindent\textbf{Experiment Results.}
Table~\ref{tab:drc_combined} reports local DRV fixing results (left) and token cost (right) across five rule categories on the 100-case dataset.
DRC-VLM-assisted methods consistently outperform the coding-agent baselines: Claude Code's overall solve rate rises from 63.0\% to 88.0\%, and Codex's from 85.0\% to 97.0\%.
Without knowledge of proprietary conditional rules, baseline agents struggle to resolve DRVs without introducing new violations; the DRC-VLM provides structured root-cause analysis, potential fixes, and connectivity impact, enabling successful repair without cascading errors.
As Figure~\ref{fig:case_study} illustrates, baseline agents attempt plausible but rule-unaware edits that trigger new violations, whereas DRC-VLM-guided agents apply rule-compliant corrections, confirming that explicit geometric reasoning is a stronger prior than iterative trial-and-error.
The remaining unsolved cases concentrate in the Color Spacing category, where tight spacing interactions risk triggering cascading violations that require net-level rerouting beyond the local repair scope.
For token cost, DRC-VLM guidance reduces average tokens by 24.5\% for Claude Code and 2.2\% for Codex, with the largest savings in Color Spacing and EN categories, reflecting fewer exploratory iterations when the agent receives rule-aware analysis upfront.

\vspace{-6pt}
\section{Conclusion and Future Works}
Local P\&R DRV fixing at advanced nodes is bottlenecked by scarce layout--violation training data and the domain gap between general-purpose VLMs and proprietary design rules.
We presented \emph{SCALE}, a framework whose self-supervised generation stage produces diverse, DRC-annotated layout--violation pairs from real BEOL context without violation labels during training.
The domain-adapted DRC-VLM provides rule-aware geometric analysis that boosts SOTA agents' solve rates by +12--25\% (up to 97\%) on 100 real sub-2\,nm cases, demonstrating that scalable, data-driven domain adaptation can bridge general-purpose VLMs to advanced-node DRC reasoning without rule-by-rule engineering.
Future work will extend to cascading multi-step DRV fixing and net-level rerouting beyond the local repair scope, as well as adapting the self-supervised generation--fine-tuning paradigm to other geometry-intensive EDA tasks where proprietary rules and data scarcity pose similar challenges.

\section*{Acknowledgment}
We gratefully thank Chen Cui, Joosung Yoon, Nithin Kumar Mali, George Kokai, Ting Ku, Yan He, Georgios Kalogerakis, and Tom Gray for their helpful discussions and support.

\newpage
\bibliographystyle{ieeetr}
\bibliography{reference}

\end{document}